\documentclass[oribib]{llncs}

\usepackage{llncsdoc}
\usepackage{amsmath}
\usepackage{amsfonts}
\usepackage{amssymb}
\usepackage{shortbold}
\usepackage{natbib}
\usepackage{float}
\usepackage{ifpdf}
\newcommand{\pdfgraphics}{\ifpdf\DeclareGraphicsExtensions{.pdf,.jpg,.png}\else\fi}
\usepackage{latexsym}
\usepackage{verbatim}
\usepackage{graphics}
\pdfgraphics
\usepackage{graphicx}
\usepackage{subfigure}
\usepackage{times}
\usepackage{color}

\newcommand{\dsum} {\displaystyle\sum}
\newcommand{\RBM}{Restricted Boltzmann Machine}
\newcommand{\KL}{KL-divergence}
\newcommand{\squishlist}{
 \begin{list}{$\bullet$}
  { \setlength{\itemsep}{0pt}
     \setlength{\parsep}{3pt}
     \setlength{\topsep}{3pt}
     \setlength{\partopsep}{0pt}
     \setlength{\leftmargin}{1.5em}
     \setlength{\labelwidth}{1em}
     \setlength{\labelsep}{0.5em} } }

\newcommand{\squishend}{
  \end{list}  }

\floatstyle{ruled}
\newfloat{algorithm}{htbp}{loa}
\floatname{algorithm}{Algorithm}

\begin{document}

\title{Comparing Probabilistic Models for Melodic Sequences}
\author{Athina Spiliopoulou \and Amos Storkey}
\institute{School of Informatics,  University of Edinburgh, United Kingdom \\
\email{\{a.spiliopoulou,a.storkey\}@ed.ac.uk}}
\maketitle

\begin{abstract}
Modelling the real world complexity of music is a challenge for machine learning. We address the task of modeling melodic sequences from the same music genre. We perform a comparative analysis of two probabilistic models; a Dirichlet Variable Length Markov Model (Dirichlet-VMM) and a Time Convolutional Restricted Boltzmann Machine (TC-RBM). We show that the TC-RBM learns descriptive music features, such as underlying chords and typical melody transitions and dynamics. We assess the models for future prediction and compare their performance to a VMM, which is the current state of the art in melody generation. We show that both models perform significantly better than the VMM, with the Dirichlet-VMM marginally outperforming the TC-RBM. Finally, we evaluate the short order statistics of the models, using the Kullback-Leibler divergence between test sequences and model samples, and show that our proposed methods match the statistics of the music genre significantly better than the VMM.
\end{abstract}


\section{Introduction}

In this paper we are interested in learning a generative model for melody directly from musical sequences. This task is challenging for machine learning methods. Repetition of musical phrases, which is essential for Western music, can occur in almost arbitrary points in time and with different degrees of variation. Furthermore, although pieces from the same genre are built using the same structural principles, the statistical relations among and within melodies from different pieces are highly complex, as melody depends on several different components, such as scale, rhythm and meter, which in many cases interdepend on each other.

Capturing the statistical regularities within a musical genre is a first step towards realistic music generation. Additionally, identifying and representing these dependencies in an unsupervised manner is particularly desirable, as descriptive features of the underlying structure of music can not only help in the analysis and synthesis of music, but also enhance the performance on a variety of musical tasks such as genre classification and music retrieval.

In this work we consider two methods for the problem of melody modeling;  a Time Convolutional Restricted Boltzmann Machine (TC-RBM) and a Dirichlet Variable Length Markov Model (Dirichlet-VMM). The first is an adaptation of the Convolutional RBM \citep{Lee2009_ConvolutionalRBM} for modeling sequential data and is motivated by the ability of RBM type models to extract high quality latent features from the input space. The second one is a non-latent variable model and is a novel form of VMM, the latter one being regarded as state of the art in melody generation \citep{Paiement2008_thesis}.

Our purpose is to answer the following questions. Are these probabilistic models able to learn the inherent structure in melodic sequences and generate samples that respect the statistics of the music genre? What aspects of the musical stucture can each of the models learn? Can melodies be decomposed into a set of musical features in the same way that images can be decomposed into sets of edges and documents into sets of topics? 

We train the models on a set of traditional reel tunes and perform a comparative analysis of these with a standard VMM. We show that the TC-RBM learns descriptive music features, such as underlying chordal structure, musical motifs and transformations of those. We assess the models on future prediction and find that our proposed methods perform significantly better than the standard VMM and are comparable to each other, with the Dirichlet-VMM having slightly higher log-likelihood. Likewise, we evaluate the short order statistics of model samples, using the Kullback-Leibler divergence, and show that samples from the TC-RBM and the Dirichlet-VMM match the statistics of the test data significantly better than samples from the VMM. 

\section{Related Work}

In many cases, the difficulties associated with modeling music have been dealt with by incorporating domain knowledge in the models. In this line of research, \citet{Paiement2008_thesis} proposes modeling different aspects of music, such as chord progressions, rhythm and melody, using graphical models and Input-Output HMMs. The structure of the models and the data representations used are based on musical theory. Additionally, \citet{Weiland2005_HHMM} propose a Hierarchical Hidden Markov Model (HHMM) for pitch. The HHMM is structurally simple and its internal states are pre-defined with respect to music assumptions.

A different course of research examines more general machine learning methods, which are able to automatically capture complex relations in sequential data, without introducing much prior knowledge. In this paper we are taking this approach and consider models that do not make assumptions explicit to music.

\citet{Lavrenko2003_MRFsPolyphonic} propose Markov Random Fields (MRFs) for modeling polyphonic music. In order for the MRF to remain tractable, much information needs to be discarded, thus making the model less suitable for realistic music.

\citet{Eck02_longtermstructureofblues} show that a Long-Short Term Memory (LSTM) Recurrent Neural Network can successfully model long-term structure in two simple musical tasks. In \citet{Eck2008_LSTM} the LSTM is extended to include meter information. The output of the network is conditioned on the current chord and specific previous time-steps, chosen according to the metrical boundaries. Trained on a set of traditional Irish reels the LSTM is shown to generate pieces that respect the reel style.

Finally, \citet{Shlomo2003_PST_Music} propose Incremental Parsing (IP) and Prediction Suffix Trees (PSTs) for modeling melodies, the latter one being the data structure used to represent VMMs. Both algorithms train simple dictionary-based predictors that parse music into a lexicon of phrases or motifs. \citet{Paiement2008_thesis} argues that despite their simple nature, these two models generate impressive musical results when sampled and can be considered state of the art in melody generation.

\section{Preliminaries}

\subsection{Musical Motifs}\label{sec:motifs}
Before describing the models, we explain the concept of motifs and their importance to music modeling, as we believe it is useful in understanding the types of structures that the VMM and the TC-RBM are trying to capture.

In Western Music, the smallest building block of a piece is called a motif. Motifs typically comprise three, four or more notes and most pieces can be expressed as a combination of different motifs and their transformations. Frequent transformations include replacement, splitting and merging of notes, and typically respect the metrical boundaries of a piece. We believe that successful capturing of music motifs can be very useful when modeling melodies, as specific motifs and their transformations are highly likely to be repeated within a piece, as well as among pieces from the same musical form.

\subsection{Variable Length Markov Model}
The VMM \citep{Ron1994_VLMMs} is a statistical model for discrete sequential data and has been shown to generate state of the art musical results when modeling melodies \citep{Shlomo2003_PST_Music}. Its advantage to a standard Markov Model (\textit{n-gram}) is that the order of the former is not fixed, but instead depends on the observed context.

A VMM is represented by a Prediction Suffix Tree. The edges of the tree are labeled with symbols from the alphabet, in this case the different music notes. Each node defines the conditional probability distribution of the next symbol given the context we acquire by concatenating all the edge symbols from the root to the node \footnote{Note that during prediction only the conditional probability distributions defined at the leaf nodes are used.}. The tree has depth $L$, but is not complete\footnote{The complete tree would represent a standard Markov Model of order $L$.}, thus giving rise to contexts that are shorter than $L$ but are still used for prediction.

To learn the tree, we start from a single root node labeled by the empty string and `grow' the tree using a breadth-first search for contexts that satisfy the following criteria: 
\squishlist
 \item The length of a context is upper bounded by a fixed length $L$
 \item The frequency counts of a context exceed a fixed threshold $c_{min}$
 \item The ratio of the conditional probability distribution defined at a node with that defined at its parent node exceeds a fixed threshold $\epsilon_{min}$
\squishend

The resulting tree comprises contexts corresponding to musical phrases that appear frequently in the data and convey significant information about the value of future time-steps. After the tree is built, the empirical conditional probability distributions are smoothed by adding a constant probability $\gamma_{min}$ to all symbols in the alphabet and renormalizing.

\subsection{Restricted Boltzmann Machine}
The \RBM\ (RBM) is a two-layer undirected graphical model with a set of visible and a set of hidden units. It is a special, bipartite form of the Boltzmann Machine \citep{Ackley1985_BMs}, in which the interaction terms are restricted to units from different layers. The joint distribution over observed and latent variables is defined through an energy function, which assigns a scalar energy to every possible configuration of the variables:
\begin{equation}
P \left( \vB,\hB|\thetaB \right) = \frac{1}{Z \left( \thetaB \right) } \exp \left( -E \left( \vB,\hB|\thetaB \right) \right) ,
\end{equation}
where $Z(\thetaB)$ is a normalizing constant called the partition function and $\thetaB$ is used to denote the set of model parameters.

In its original form, an RBM has binary, logistic units in both layers\footnote{However, see for example \citet{Welling_exponential_harmoniums} on how to define RBMs with real-valued units.} and its energy function is defined as:
\begin{equation}
 E \left( \vB,\hB|\thetaB \right) = -\cB^{\sf T} \vB - \bB^{\sf T} \hB - \vB^{\sf T} \WB \hB ,
\end{equation}
where $\cB$ and $\bB$ are the biases for the visible and hidden units, respectively, and $\WB$ is the weight matrix for the interaction terms.

Inference in this model can be performed efficiently using block Gibbs sampling, as due to the bipartite structure of the model, the conditional distributions of the hidden units given the visibles and of the visible units given the hiddens factorize.

Maximum Likelihood learning in the RBM is difficult due to the partition function $Z(\thetaB)$ which is typically intractable\footnote{Computing the partition function involves a sum over all possible configurations of visible and hidden units.}. However, parameter estimation can be performed using Contrastive Divergence \citep{Hinton2002_PoE_CD}, an objective that approximates the likelihood and has been shown to work well in practice.

\section{Models}\label{sec:models}

\subsection{Dirichlet-VMM}\label{sec:Dirichlet-VMM}
The VMM is similar to an n-gram model in that its performance is significantly influenced by the smoothing technique used. An alternative to a standard form of variable length Markov model is a hierarchical model, where each conditional multinomial distribution in the tree is sampled from a dirichlet Distribution, centered at the sample multinomial for the parent node. In this model smoothing is performed implicitly by taking a Bayesian approach and introducing an appropriate prior distribution at each node while building the tree.

More formally, let $\mB \left( \xB_t|\xB_{t-1},\ldots,\xB_{t-\tau-1} \right)$ be defined by 
\[
 m_k = P \left( \xB_t = k|\xB_{t-1},\ldots,\xB_{t-\tau-1} \right) \enspace .
\]
Then we model each conditional distribution as: 
\begin{equation}
P \left( \xB_t|\xB_{t-1},\ldots,\xB_{t-\tau} \right) \sim \mbox{Dirichlet} \left( \alpha  \mB \left( \xB_t|\xB_{t-1},\ldots,\xB_{t-\tau-1} \right) \right) \enspace .
\end{equation}
This forms a hierarchical tree with the marginal distribution $P(\xB_t)$ as the root node, and successively more specific conditional distributions as we traverse down the tree. The intermediate nodes, though identified with particular distribution, are not used directly to model the data; that is done by the leaf nodes. 

Learning this hierarchical distribution involves learning the posterior distributions at each level of the hierarchy from the data associated with the given node (i.e. the data that satisfies the conditional distribution).
\begin{align}
 P (&\xB_t|\xB_{t-1},\ldots,\xB_{t-\tau},D) \sim \nonumber \\
 & \mbox{Dirichlet} ( \alpha E \left[ \mB(\xB_t|\xB_{t-1},\ldots,\xB_{t-\tau-1},D) \right] + \cB (\xB_t,\xB_{t-1},\ldots,\xB_{t-\tau}) ) ,
\end{align}
where the $c_k(\xB)$ function counts the number of occurrences of sequence $\xB$ in the dataset where the last element is in state $k$, and $E$ denotes expectation.

The mean of the posterior Dirichlet at each node is the prior Dirichlet for the data at the child nodes. Note the top levels of the hierarchy have a large amount of associated data, but as we progress down the tree the amount of data reduces. In the limit where there is no data the posterior distribution for that node is just given by the posterior for the parent node.

This model is directly related to the sequence memoizer \citep{Wood09_memoizer}, but is a finite model using Dirichlet distributions, instead of a Pitman Yor model. Using Dirichlet distributions makes the inference procedure entirely conjugate and thus no sampling is required. We call this model a Dirichlet-VMM in this paper.

\subsection{Time Convolutional RBM} 
 \begin{figure}[t]
  \centering
  \includegraphics[scale=0.15]{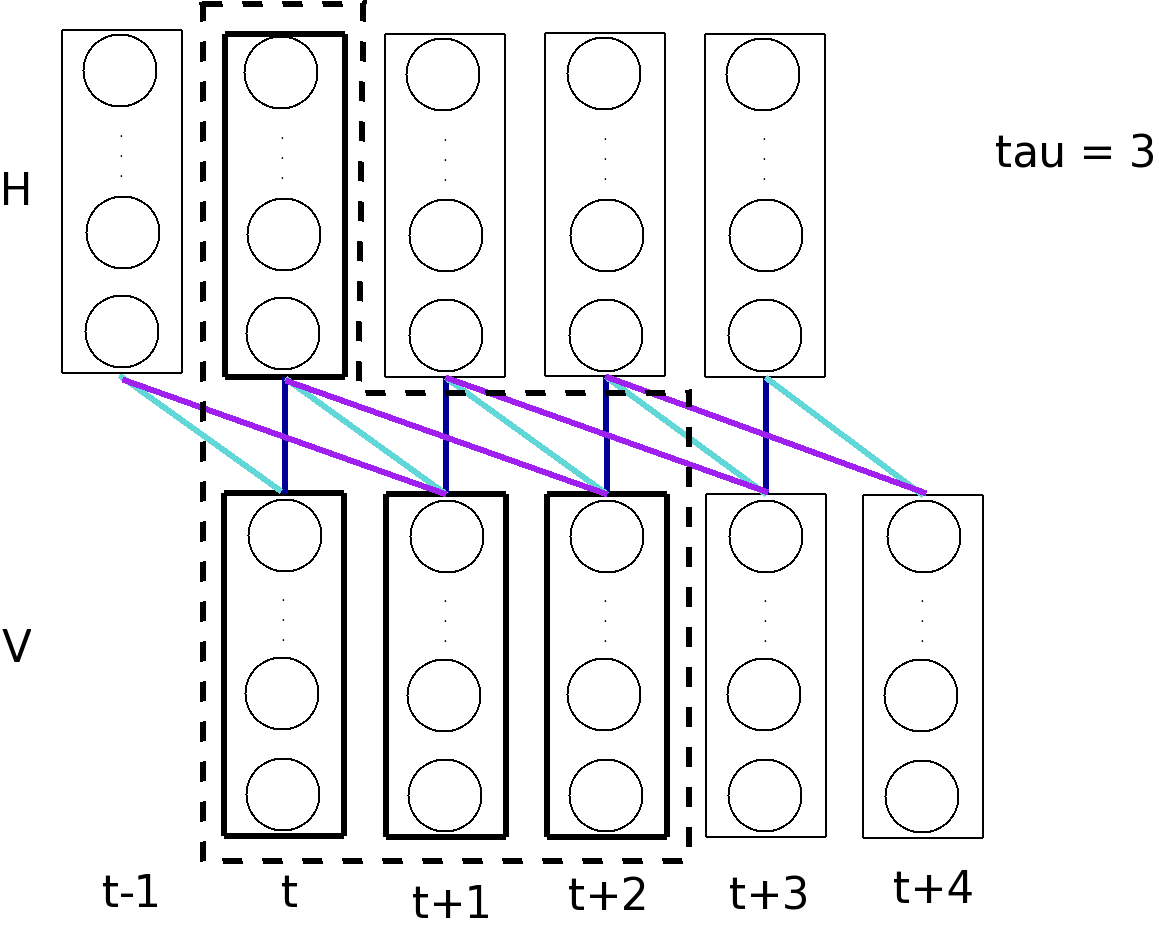}
  \caption[]{A Time Convolutional RBM with filter size $\tau=3$. The dashed frame shows the connections to the hidden units in a single time-step. Each unit receives input from all visible units in a subsequence of length $\tau$. The model is `unrolled' in time through a weight sharing mechanism.}
  \label{fig:TCRBM}
 \end{figure}
 
We propose a Time Convolutional RBM (TC-RBM) as a new way of modeling sequential data with an RBM type network. We believe that models based on the RBM are particularly suitable for capturing the componential structure of music, as they can learn distributed representations of the input space, decoupling the different factors of variation into features being ``on'' or ``off''. The TC-RBM is an adaptation of the Convolutional RBM for sequences and it is motivated by the successful application of such models in static image data \citep{Lee2009_ConvolutionalRBM, Norouzi09_CRBMs}.

Previous RBM approaches to sequence modeling use the RBM to model a single time-step and attempt to capture the temporal relations in the data by introducing different types of directed connections from units in previous time-steps \citep{Taylor2006_CRBM, Sutskever2007_TRBM, Taylor2009_FactoredCRBM}. In contrast, the TC-RBM is a fully undirected network and attempts to capture the structure of music at a motif level rather than a single time-step.

The TC-RBM is depicted in Fig.\ \ref{fig:TCRBM}. Local temporal dependencies are captured by learning an RBM on visible subsequences of fixed length - instead of single data points. This allows the hidden units to learn valid configurations for a whole subsequence and thus capture frequent motifs and their transformations. Longer sequences are modelled by applying convolution through time. This weight sharing mechanism allows us to better model boundary effects and provides the model with translation invariance along time, which is desirable as motifs can appear anywhere in a musical piece.

The energy function of the TC-RBM is defined as:
\begin{equation}
 E \left( \VB,\HB|\thetaB \right)= -\dsum_t \left( \cB^{\sf T} \vB_t + \bB^{\sf T} \hB_t + \dsum_{k=0}^{\tau-1} \vB_{t+k}^{\sf T} \WB_{k} \hB_t \right) ,
\end{equation}
where $\VB$ is a visible sequence, $\HB$ is the hidden configuration for that sequence and $\tau$ is the size of the filter we apply\footnote{The filter size is the number of visible time-steps that a hidden unit receives input from.}. The interaction terms are parameterized by the weight tensor\footnote{Each slice $k$ of the $\WB$ tensor is the weight matrix for the connections of hidden units at time $t$ with the visible units at time $t+k$.} $\WB$ and the unit biases, $\cB$ and $\bB$ for visible and hidden units respectively, are the same for all time-steps.

Similarly to an RBM, the joint probability distribution of the observed and hidden sequence under the TC-RBM is defined as $P(\VB,\HB|\thetaB) = \exp \left( - E(\VB,\HB|\thetaB) \right) / Z(\thetaB)$.

The conditional probability distributions of this model factorize over time and units and are given by softmax and logistic functions:
\begin{equation}
 P(v_{i,t}=1| \HB) = \frac{\exp \left( c_i + \sum_{k=0}^{\tau-1} \WB_{i,\cdotp,k} \hB_{t-k} \right) }{\dsum_q \exp \left( c_q + \sum_{k=0}^{\tau-1}\WB_{q,\cdotp,k} \hB_{t-k} \right) },
\label{eq:cond1}
\end{equation}
\begin{equation}
 P(h_{j,t}=1|\VB) = \left[ 1+\exp \left( -b_j - \!\dsum_{k=0}^{\tau-1}\!\vB_{t+k}^{\sf T} \WB_{\cdotp,j,k} \right) \right]^{-1} \enspace .
 \label{eq:cond2}
\end{equation}

Inference can be performed using block Gibbs sampling. The computation of (\ref{eq:cond1}) and (\ref{eq:cond2}) can be performed efficiently by convolving along the time dimension the appropriate slice of the weight tensor with the hidden and visible sequence respectively. As in the RBM, learning can be performed using the Contrastive Divergence rule.

\section{Experiments}\label{sec:experiments}

In the following section we want to assess the ability of the models to learn the inherent structure of melodic sequences belonging to the same genre. An appropriate measure for this evaluation is the marginal likelihood of the data under each model $M$, $P(\DB|M)=\int{P(\DB|\thetaB,M)P(\thetaB|M)d\thetaB}$. However, computing the marginal likelihood under the TC-RBM is intractable\footnote{Computing the data likelihood $P(\DB|\thetaB,M)$ involves a sum over all possible configurations of visible and hidden units.} and thus we need to make use of other quantitative measures.

In the music modeling literature, evaluation is primarily based on qualitative analysis, like listening to model generations. To our knowledge, the only quantitative measures used so far are next-step prediction accuracy \citep{Paiement2008_thesis, Lavrenko2003_MRFsPolyphonic} and perplexity \citep{Lavrenko2003_MRFsPolyphonic}. In this work, we broaden this evaluation framework to consider longer future prediction, instead of only next-step, as this provides an insight regarding model performance through time.

To make our comparative analysis more rigorous, we also examine the short order statistics of the models and compare them with the data statistics. To perform this analysis we compute the Kullback-Leibler divergence between the frequency distribution of events in test sequences and in model samples, which measures how well the model statistics match the data, or put differently, how much a model has yet to learn.

Besides the quantitative evaluation, we are also interested in assessing the capabilities of the models to identify and represent the statistical regularities of the data. In the VMM models, the learned lexicon of phrases determines the frequent musical motifs, but does not provide any information regarding the underlying structure, as the encoded patterns are fixed. On the other hand, the TC-RBM learns a distributed representation of the input space; a set of latent features that are `on' or `off' depending on the input signal. We demonstrate that these features are music descriptors extracted from the data and convey information regarding music components such as scale, octaves and chords.

\subsection{Data Processing and Representation}\label{sec:representation}

In the following experiments we use a dataset comprising 117 traditional reels collected from the Nottingham Folk
Music Database\footnote{We use the MIDI toolbox \citep{Eerola2004} to read and write MIDI files.}. Reels are traditional Scottish and Irish tunes used to accompany dances. All tunes are in the G major scale and have 4/4 meter.

\begin{figure}[t]
 \centering
 \includegraphics[scale=0.35]{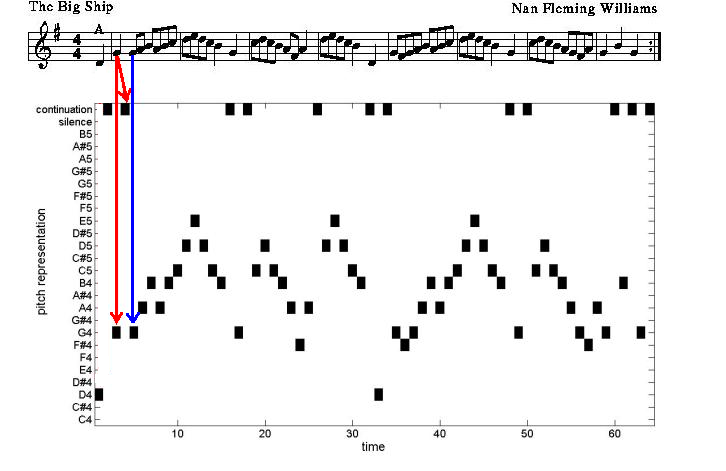}
 \caption{Data Representation: Time is discretized in eighth note intervals. At each time-step, pitch is represented by a $1$-of-$26$ dimensional vector. Red (left) arrows: a $G4$ quarter note lasts for two time-steps and is represented by $G4$ followed by `continuation'. Blue (right) arrow: a $G4$ eighth note lasts for one time-step and is represented by a single $G4$.}
 \label{fig:data}
\end{figure}

Our representation is depicted in Fig.\ \ref{fig:data}. The components we wish to model are pitch and duration of the notes in the melody. Duration is modelled implicitly by discretizing time in eighth-note intervals. At each time-step, pitch is encoded using a $1$-of-$m$ vector. We use only two octaves, $C4$-$B5$, giving rise to a $24$-dimensional vector. Values outside this octave range are trunctated to the nearest octave.

Finally, we augment the $1$-of-$m$ vector with two more values. The first one is used to represent music silence. The second one is used to represent `continuation' of an event and allows us to keep more accurate information concerning the duration of notes. 

\subsection{Implementation Details }
We trained a VMM, a Dirichlet-VMM and a TC-RBM. To set the parameters $c_{min}$, $\epsilon_{min}$ and $\gamma_{min}$ of the VMM we applied grid search over the product space of the parameters and chose the values that maximize the data log-likelihood using leave-one-out cross validation on the training data. We used the same grid search procedure to set the parameter $\alpha$ of the Dirichlet prior in the Dirichlet-VMM\footnote{In the VMM, the maximum length $L$ was set to a very large value ($100$), which resulted in the depth of the tree being controlled by the parameter $c_{min}$ for the frequency counts. The resulting depth for the optimal tree is 13. In the Dirichlet-VMM, we used a global $\alpha$ parameter and applied grid search over the product space of $c_{min}$, $\epsilon_{min}$ and $\alpha$.}.

For the TC-RBM, we used 50 hidden units. We chose the size of the filters to be 8 time-steps, which corresponds to the length of a music bar. For learning the model we used the following settings: CD-5, 500 epochs, 0.5  learning rate decreasing on a fixed schedule, 0.0002 weight decay. We additionally used a sparsity term\footnote{It has been suggested \citep{Lee2009_ConvolutionalRBM, Norouzi09_CRBMs} that due to the over-complete hidden representation of convolutional RBMs, encouraging sparsity is important and can facilitate learning.} in the objective function, which encourages hidden units to be `off'. We implemented sparsity as described in \citet{Lee2007_V2}, and set the desired activity level to 0.1. 

\subsection{Learning Musical Features}
\begin{figure*}[t]
 \includegraphics[width=\linewidth, scale=0.5]{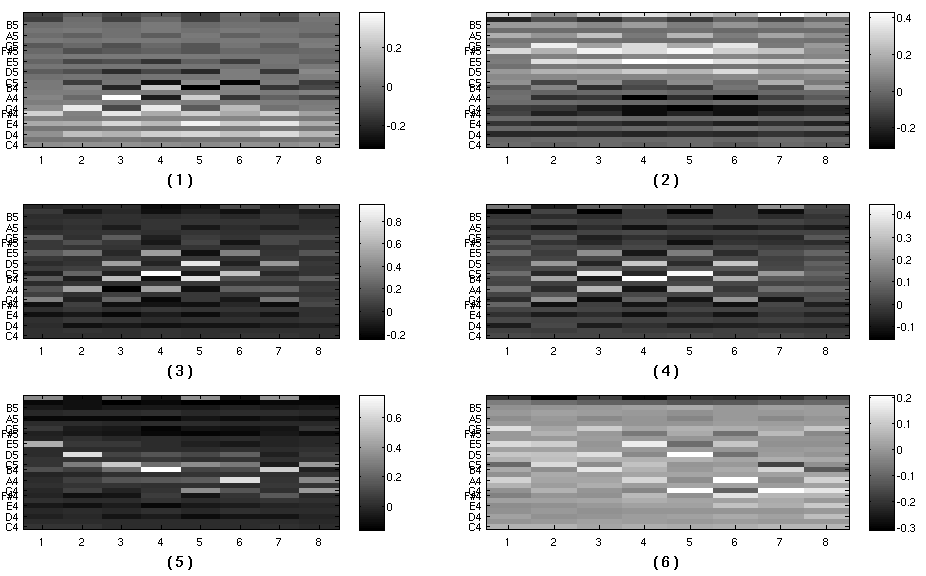}
 \caption{Weight filters for 6 different hidden units of the learned TC-RBM. All units prefer notes from the G major scale to be `on'; these notes are explicitly ticked in the $y$-axis. Filters 1 and 2 respond to similar patterns, but operate in the lower and higher octave respectively. Filters 3 and 4 respond to notes from either the $Gmaj$ or the $Am$ chord in alternate time-steps. Filter 5 is highly selective to a specific motif, whereas filter 6 responds to several configurations of the scale notes.}
 \label{fig:filters}
\end{figure*}

In the TC-RBM each hidden unit is connected with all the visible units from eight subsequent time-steps. This gives rise to a $26 \! \! \times \! \! 8$-dimensional filter for each hidden unit\footnote{The filter for hidden unit $j$ is the slice $\WB_{\cdotp,j,\cdotp}$ of the weight tensor.}. 

The filters corresponding to 6 different hidden units from the learned TC-RBM are depicted in Fig.\ \ref{fig:filters}. We can notice that all units prefer visible configurations with notes from the G major scale\footnote{Notes from the G major scale: $G A B C D E F^\# G$} to be `on', but have various degrees of selectivity and respond to different subsets of these notes in different positions. 

For instance, filter (6) is fairly broad and may respond to several different configurations of notes from the G major scale, whereas filter (5) is highly selective, responding primarily to the downwards-upwards movement $EDCBGAB$ through the scale and certain variations of it.

An interesting property of the top two filters is their relation with respect to the octave. Both units respond to similar music phrases. For instance, both units respond to the motif $F^\#GAB$ starting at either position 1 or 3. However, the left unit operates in the lower octave ($C4$-$B4$), whereas the right one operates in the higher octave ($C5$-$B5$). 

Another interesting property is the relation of the filters to the tone chords of the scale\footnote{These are chords of three, four or five notes built from alternate scale notes of $G$ Major.}. In several filters, the prefered subset of notes at each time-step corresponds to the notes of a tone chord. This property is particularly prominent in filters (3) and (4). For instance in filter (3), the prefered subset of notes at odd time-steps corresponds to the notes of the $Gmaj$ chord ($GBD$), whereas at even time-steps to the notes of the $Am$ chord ($ACE$).

\begin{figure*}[t]
 \includegraphics[width=\linewidth]{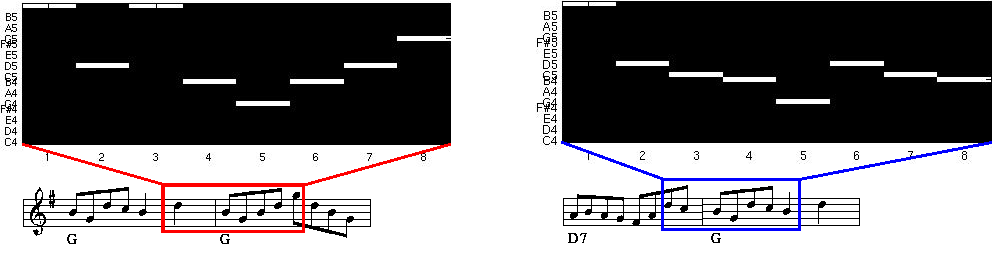}
 \caption{Two different visible configurations that frequently turn `on' the hidden unit corresponding to filter (5) from Figure \ref{fig:filters} during sampling. The visible configurations are also depicted in the musical score. Each configuration contains a different variation of the motif $D\!*\!BG$, which is prominent in filter (5).}  
 \label{fig:visibles}
\end{figure*}

In order to better understand how the filters behave, we looked at random visible configurations that tend to activate a hidden unit during sampling. Figure \ref{fig:visibles} shows two such visible configurations for the hidden unit corresponding to filter (5). Although the two configurations seem fairly different, they both contain the motif $D \!* \!BG$ in positions $2$ to $5$ with either a pass through $A$ or `continuation' of $D$ in position 3. Filter (5) is highly responsive to this motif, and although time-steps 6 to 8 in the visible configurations are not highly preferable, the unit is still very likely to turn `on'.

Overall, we can see that the learned filters encode familiar musical movements, such as arpeggios and scales\footnote{These can be loosely defined as groups of subsequent scale notes, either going up or going down.}. However, the interesting and potentially powerful characteristic of the TC-RBM representation is that it also encodes and groups together many possible transformations of these movements. Therefore, in contrast to the VMM representation, the motifs learned by the TC-RBM are not fixed; the TC-RBM filters group together musically sound variations of motifs, thus encoding possible note substitutions, merging and splitting. This is very advantageous when modeling music, given its highly complex and ingenious nature, and also allows for more genuine music generations.

\subsection{Prediction Task}
Given an observed test subsequence we want to evaluate how well a model can predict the following $k$ time-steps. We define the prediction log-likelihood of a test sequence $D$ under a model $M$ with parameters $\thetaB$, as the log probability of the actual future time-step $d_{t+\tau}$ given time-steps $d_1$ up to $d_t$, averaged over all time-steps $T_n$ of the test sequence. More specifically: 
{\setlength\abovedisplayskip{3pt plus 2pt minus 0.5pt}
\setlength\belowdisplayskip{2pt plus 2pt minus 0.5pt}
\begin{equation}
 \log L_{\tau}(\thetaB, M; \DB) = \frac{1}{T_n} \dsum_t^T \log P(d_{t+\tau}|d_1,\ldots,d_t, \thetaB, M) \enspace .
 \label{eq:predictiveDistribution}
\end{equation}}
We use the empirical marginal distribution\footnote{The empirical distribution of the training data under the assumption that all time-steps are iid (independently and identically distributed). This distribution is the best predictor in the absence of temporal dependencies.} as a baseline for evaluating model performance.

\subsubsection{Computing Prediction $\log L$ under the VMM and the Dirichlet-VMM.}
For $\tau=1$ we can compute (\ref{eq:predictiveDistribution}) exactly under the VMM models. For $\tau > 1$ we need to marginalize over the future time-steps that are between $d_t$ and $d_{t +\tau}$, ie:
\begin{equation}
 P \left( d_{t+\tau}|d_1,\ldots,d_t, \thetaB, M \right) = \! \! \! \! \! 
 \dsum_{d_{t+1},\ldots, d_{t+\tau-1}} \! \! \! \! \! \! \! P(d_{t+1}, \ldots, d_{t+\tau}|d_1,.\,.\,.\,,d_t, \thetaB, M) \enspace .
\end{equation}

We approximate this distribution by drawing a number of sampled paths from the VMM and averaging over the conditional probability distributions defined by these paths which are given exactly under the VMM:
\begin{equation}
 \! \! \! P(d_{t+\tau}|d_1,\ldots,d_t, \thetaB, M) \approx
 \frac{1}{S} \dsum_{s = 1}^S \! P(d_{t+\tau}| d_1,\ldots,d_t, d_{t+1}^s,\ldots,d_{t+\tau-1}^s, \thetaB, M).
\end{equation}
We use 100 sampled paths in the experiments reported here.

\subsubsection{Computing Prediction $\log L$ under the TC-RBM.}
 In order to evaluate (\ref{eq:predictiveDistribution}) under the TC-RBM, we need to marginalize over future visible time-steps that are between $d_t$ and $d_{t +\tau}$ for $\tau\! >\! 1$ and over the possible configurations of hidden units for time-steps $t$ to $t\!+\!\tau$. To avoid this computation we approximate the predictive distribution using samples from the model. The sampling procedure is given in Algorithm 1.

\begin{algorithm}[h]

\vspace{2pt}
Let $\VB$ be a visible sequence and $\HB$ a hidden sequence

\vspace{2pt}
Initialize $\VB$ randomly

\vspace{2pt}
Set $\vB_{1...t} = d_{1...t}^n$

\vspace{2pt}
While $s <$ numberOfSamples

\vspace{2pt}
$ \quad \quad \HB \sim P_{TC-RBM}(\HB|\VB, \thetaB) \quad$ (Equation 7)
   
$ \quad \quad \VB \sim P_{TC-RBM}(\VB|\HB, \thetaB) \quad$ (Equation 6)
   
$ \quad \quad$Clamp to context: $\vB_{1...t} = d_{1...t}^n$

\vspace{2pt}
$ \quad \quad$If equilibrium reached

$ \quad \quad \quad \quad \HB^s = \HB$
   
$ \quad \quad \quad \quad s = s + 1$
   
$ \quad \quad$end

end

\vspace{2pt}
$P(d_{t+\tau}^n|d_{1...t}^n, \thetaB, M) = \frac{1}{S} \dsum P_{TC-RBM}(\vB_{t+\tau}|\HB^s, \thetaB)$

\label{alg:sampling}
\caption{Sampling Procedure for the TC-RBM}
\end{algorithm}
 
 In our experiments, we use 100 chains and run 15 Gibbs iterations within each chain. Overall, we use 500 samples to approximate the predictive distribution, discarding the first 10 samples from each chain.

\subsubsection{Results.}

\begin{figure}[t]
\includegraphics[scale=0.3, trim=0 0 0 2cm]{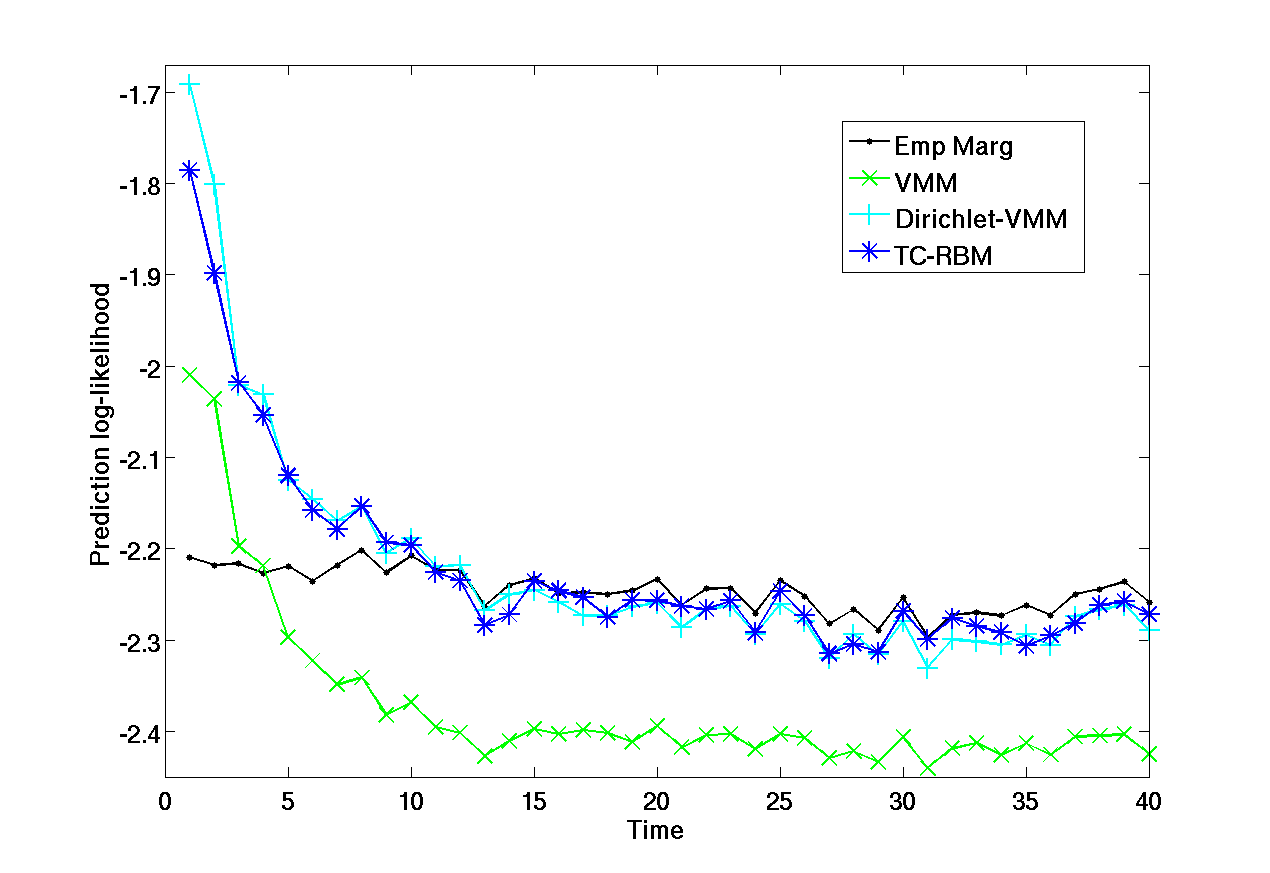}
\caption{Prediction Log-Likelihood under different models plotted as a function of time. The Log-Likelihood is averaged across 2,000 configurations of context-future observations, randomly selected from the test data.}
\label{predLogL}
\end{figure}

Figure \ref{predLogL} shows the log-likelihood of predicting the true succession given an observed sub-sequence from a test tune under different models. As already mentioned, our baseline for assessing model performance is the empirical marginal distribution. The log-likelihood of the test data under the empirical marginal corresponds to the black curve.

Compared to the empirical marginal distribution, the standard VMM (green x) performs significantly better in predicting the first two future time-steps, only slightly better for time-steps 3 and 4 and significantly worse than the empirical marginal after the 5th time-step. 

Both the Dirichlet-VMM (cyan crosses) and the TC-RBM (blue stars) perform significantly better than the standrad VMM in predicting all future time-steps. These two models have similar performance in the prediction task, with the Dirichlet-VMM outperforming the TC-RBM for the first two time-steps and their prediction log-likelihood being almost the same from the 3rd time-step onwards. 

We should note that the performance of the TC-RBM in prediction may be compromised by the fact that the block Gibbs procedure samples the future sub-sequence as a whole at each iteration. This means that due to the convolutional structure of the model, the time-step we are trying to predict receives information not only from the past, which is clamped to the observed context, but also from the future which is initialized randomly and can thus drive the samples into different energy basins. 

Compared to the empirical marginal distribution, both the Dirichlet-VMM and the TC-RBM perform better for the first 10 time-steps. The prediction log-likelihood under the models is initially much higher than the one under the empirical marginal distribution, but decays as we try to predict further into the future. The models slowly forget the information upon which they have been conditioned and after the 10th time-step converge to a steady-state distribution, which is slightly worse than the empirical marginal distribution for prediction. 

While long-term prediction is useful for characterizing model behaviour through time, it is not adequate for evaluating the generative capabilities of the models. For instance, even if a musical phrase is highly predictable given a certain context, the models can get bad predictive performance if they are not able to determine the correct starting time-step for the phrase. 

Nevertheless, we can note that in contrast to the standard VMM, our proposed models converge to the empirical marginal distribution over time and thus are better in capturing the statistical regularities in the data, which is the first step towards realistic music generation.

\subsection{Using the Kullback-Leibler divergence to compare statistics}
The Kullback-Leibler (KL) divergence is a measure of how different two probability distributions, P and Q, are. For discrete random variables, it is defined as $D_{\mathrm{KL}}(P\|Q) = \sum_i P(i) \log \frac{P(i)}{Q(i)}\!$ and shows the average number of extra bits needed to encode events from a distribution P with a code based on an approximating distribution Q. If the true distribution that generated the data is P and the model distribution is Q, then the lower the KL-divergence the better the model matches the data.

To compare model statistics with data statistics, we compute the frequency distribution of events in samples generated by each of the models and in test sequences, and compute the KL-divergence between the normalized data and model frequencies. More specifically, let $d_t$ denote the observation of a single time-step at time $t$. Then to compare first-order statistics we estimate the KL-divergence between $P(d_t)$ and $Q(d_t)$ by computing:
\begin{equation}
 D_{\mathrm{KL}}(P_{\mathrm{data}}(d_t)\|Q_M(d_t)) = \frac{1}{N}\dsum_{n=1}^N P_{\mathrm{data}}(d_t^n) \log \frac{P_{\mathrm{data}}(d_t^n)}{Q_M(d_t^n)},
\end{equation}
where $P_{\mathrm{data}}(d_t)$ is the empirical marginal distribution of data sequences and $Q_M(d_t)$ is the marginal distribution of samples generated by model $M$. Similarly, for pairwise statistics we compute $D_{\mathrm{KL}}(P_{\mathrm{data}}(d_t,d_{t+1})\|Q_M(d_t,d_{t+1}))$, for third order statistics $D_{\mathrm{KL}} (P_{\mathrm{data}}(d_t,d_{t+1},d_{t+2})\|Q_M(d_t,d_{t+1},d_{t+2}))$, and so on. 

Since the true distribution that generated the data is unknown, we perform a bootstrapping procedure for the estimation of the KL-divergence. More specifically, we compute the KL-divergence for each statistic $50$ times, each time using a different data resample, obtained by random sampling with replacement from the original test dataset. In our results, we report the mean and variance of the KL-divergence for each statistic.

The number of possible events grows exponentially with the order we consider, which makes the statistics for higher-orders less reliable, given that we have a finite test set. In order to get a better understanding of how the models perform through time, we additionally consider pairwise statistics with lags, that is statistics of events comprising two time-steps which are not adjacent in time. For instance for lag $l=1$ we consider the frequencies of events $ \left(d_t,d_{t+2} \right) $, for lag $l=2$ we consider $\left(d_t,d_{t+3}\right)$ and so on.

\begin{table}[t]
\caption{KL-divergence between data statistics and model statistics. We report the mean and variance (in the parenthesis) of the KL-divergence for each statistic. These are computed using 50 resamples, obtained by random sampling with replacement from the test dataset.}
\begin{tabular}{|c||c|c|c|c|c|c|c|}
\hline
& order $1$ & order $2$ & order $3$ & order $4$ & order $5$ & order $6$ \\
\hline
Trainset & $0.032\,(1e$-$4)$ & $0.433\,(0.012)$  & $1.351\,(0.093)$ & $3.150\,(0.197)$ & $5.455\,(0.330)$ & $7.985\,(0.479)$ \\
TC-RBM   & $0.064\,(2e$-$4)$ & $0.273\,(4e$-$4)$ & $0.872\,(0.002)$ & $2.420\,(0.047)$ & $5.244\,(0.248)$ & $8.584\,(0.645)$ \\
Dir-VMM  & $0.045\,(3e$-$4)$ & $0.302\,(0.005)$  & $1.158\,(0.076)$ & $2.594\,(0.172)$ & $4.295\,(0.357)$ & $6.462\,(0.672)$ \\
VMM     & $0.187\,(1e$-$4)$ & $0.481\,(4e$-$4)$ & $1.331\,(0.023)$ & $3.242\,(0.114)$ & $5.839\,(0.284)$ & $8.772\,(0.452)$ \\
\hline
& lag $1$ & lag $2$ & lag $3$ & lag $4$ & lag $5$ & lag $6$ \\
\hline
Trainset & $0.228\,(0.002)$  & $0.251\,(0.003)$  & $0.188\,(8e$-$4)$ & $0.236\,(0.003)$  & $0.180\,(8e$-$4)$ & $0.254\,(0.001)$  \\
TC-RBM    & $0.229\,(6e$-$4)$ & $0.203\,(5e$-$4)$ & $0.222\,(6e$-$4)$ & $0.201\,(5e$-$4)$ & $0.204\,(9e$-$4)$ & $0.175\,(4e$-$4)$ \\
Dir-VMM  & $0.198\,(0.001)$  & $0.175\,(0.001)$  & $0.224\,(0.001)$  & $0.184\,(0.002)$  & $0.215\,(0.001)$  & $0.202\,(0.001)$  \\
VMM     & $0.476\,(2e$-$4)$ & $0.474\,(4e$-$4)$ & $0.542\,(4e$-$4)$ & $0.477\,(6e$-$4)$ & $0.533\,(5e$-$4)$ & $0.472\,(6e$-$4)$ \\
\hline
\end{tabular}
\end{table}

\subsubsection{Results.} Table 1 shows the mean and variance of the \KL\ between the statistics of test sequences and a priori samples for various models. The first row compares test sequences to train sequences and is used as a reference for interpreting the results. Looking at the first order statistics we can note that the TC-RBM and the Dirichlet-VMM have much lower \KL\ than the VMM, with the Dirichlet-VMM having the lowest amongst the models. In fact the \KL\ for the former two models is very close to the \KL\ between test sequences and train sequences, which indicates that samples generated from these models match the statistics of the test data well.

For the second, third and fourth order statistics, the TC-RBM has the lowest \KL, with the Dirichlet-VMM following closely and the VMM lagging behind. Interestingly, the \KL\ of these statistics for the TC-RBM and the Dirichlet-VMM is even lower than the one for the train data. We believe that this stems from the fact that the models are capturing the underlying structure that characterizes the whole musical genre, and to some extent ignore the finer structure that characterizes each individual music piece. This can result in model samples that have higher inter- and lower intra-piece similarity than a set of real music sequences.

For fifth and sixth order statistics, the \KL\ for the TC-RBM and the VMM is close to the \KL\ for the train data, whereas for the Dirichlet-VMM is lower. As mentioned earlier, the estimates for higher order statistics are less reliable, since the number of possible configurations is exponentially large and thus very difficult to characterize from a finite set of samples. 

Finally, for the pairwise statistics with lags, the \KL\ for both the TC-RBM and the Dirichlet-VMM is low, very close to the one for the train data, whereas for the VMM it is considerably higher. This suggests that our proposed methods respect the short order statistics of the musical genre and are better than the VMM in capturing the statistical regularities of the data through time.

\section{Discussion}

We addressed the problem of learning a generative model for music melody by considering two probabilistic models, the Dirichlet-VMM and the Time Convolutional RBM. We showed that the TC-RBM, trained on a dataset of tunes from the same genre, learns descriptive musical features that can be used to decompose the underlying structure of the data into musical components such as scale, octave and chord. 

We performed a comparative analysis of the two models with the standard VMM, which, to our knowledge is state of the art in melody generation. We showed that in a long-term prediction task both models perform significantly better than the VMM and comparably with each other. The Dirichlet-VMM is a better next-step predictor, which can be partially accredited to its main strength, that is its ability to use shorter or longer contexts depending on whether they provide useful information or not. 

We evaluated the short order statistics of the models by comparing the Kullback-Leibler divergence between test sequences and model samples. We demonstrated that sampled generations from our proposed methods match the statistics of the test sequences considerably better than samples from the VMM and respect the genre statistics, as the \KL\ for the TC-RBM and the Dirichlet-VMM is very close to the \KL\ between test and train sequences.

The ability of the TC-RBM to extract descriptive musical features allows us to consider hierarchical approaches for melody generation, which can help modulate the appearance of features through time. We are currently experimenting with deeper TC-RBM architectures, where TC-RBMs are stacked on top of one another in a greedy manner (see \citet{Hinton2006_CD_DBNs} for the RBM case). Deep models have been shown to learn hierarchical representations of the input space, where more abstract features are captured in higher layers, which according to the tonal music theory \citep{Lerdahl1983_generativeTheoryMusic} is how music composition should be understood.

Finally, an interesting direction for future research in music modeling involves exploration of methods that can distinguish between inter- and intra-piece similarity. The methods examinded in this work can learn the statistical relations within a musical genre, but are not able to effectively model piece-wise variation. Considering methods that enable us to sample a prior distribution for each piece, such as topic models, would be a first step towards this direction.

\subsubsection*{Acknowledgements.} Athina Spiliopoulou is partly funded by an EPSRC scholarship.
\bibliographystyle{apalike}

\begin{thebibliography}{References}
{\small
\vspace{-10pt}

\bibitem[Ackley et~al., 1985]{Ackley1985_BMs}
Ackley, D.H., Hinton, G.E., Sejnowski, T.J. (1985). A learning algorithm for
  {B}oltzmann machines. Cognitive Science  9(1), 147--169.

\bibitem[Dubnov et~al., 2003]{Shlomo2003_PST_Music}
Dubnov, S., Assayag, G., Lartillot, O., and Bejerano, G. (2003). Using machine-learning methods for musical style modeling. Computer 36(10), 73--80.

\bibitem[Eck and Lapalme, 2008]{Eck2008_LSTM}
Eck, D. and Lapalme, J. (2008). Learning musical structure directly from sequences of music. Technical report, Universit\'{e} de Montreal.

\bibitem[Eck and Schmidhuber, 2002]{Eck02_longtermstructureofblues}
Eck, D. and Schmidhuber, J. (2002). Learning the long-term structure of the blues. In: Dorronsoro, J.R. (ed.)
ICANN. LNCS, vol. 2415, pp. 284--289. Springer.

\bibitem[Eerola and Toiviainen, 2004]{Eerola2004}
Eerola, T. and Toiviainen, P. (2004). MIDI Toolbox: MATLAB Tools for Music Research. University of Jyv{\"a}skyl{\"a}, Jyv{\"a}skyl{\"a}, Finland, \url{www.jyu.fi/musica/miditoolbox/}

\bibitem[Hinton, 2002]{Hinton2002_PoE_CD}
Hinton, G.E. (2002). Training products of experts by minimizing contrastive divergence. Neural Computation 14(8), 1771--1800.

\bibitem[Hinton et~al., 2006]{Hinton2006_CD_DBNs}
Hinton, G.E., Osindero, S., and Teh, Y.W. (2006). A fast learning algorithm for deep belief nets. Neural Computation 18(7), 1527--1554.

\bibitem[Lavrenko and Pickens, 2003]{Lavrenko2003_MRFsPolyphonic}
Lavrenko, V. and Pickens, J. (2003). Polyphonic music modeling with random fields. In: Rowe, L.A., Vin, H.M., Plagemann, T., Shenoy, P.J., Smith, J.R. (eds) ACM Multimedia. Proceedings of the Eleventh ACM International Conference on Multimedia, p. 120--129. ACM.

\bibitem[Lee et~al., 2008]{Lee2007_V2}
Lee, H., Ekanadham, C., and Ng, A.~Y. (2008). Sparse deep belief net model for visual area {V2}. In: Platt, J.C., Koller, D., Singer, Y., Roweis, S.T. (eds.) NIPS. Advances in NIPS 20. MIT Press.

\bibitem[Lee et~al., 2009]{Lee2009_ConvolutionalRBM}
Lee, H., Grosse, R., Ranganath, R., and Ng, A.Y. (2009). Convolutional deep belief networks for scalable unsupervised learning of hierarchical representations. In: Danyluk, A.P., Bottou, L., Littman, M.L. (eds.) ICML. ACM ICPS, vol. 382, p. 77. ACM.

\bibitem[Lerdahl and Jackendoff, 1983]{Lerdahl1983_generativeTheoryMusic}
Lerdahl, F. and Jackendoff, R. (1983). A Generative Theory of Tonal Music. The MIT Press, Cambridge, Massachusetts, London, England.

\bibitem[Norouzi et~al., 2009]{Norouzi09_CRBMs}
Norouzi, M., Ranjbar, M., and Mori, G. (2009). Stacks of convolutional restricted {B}oltzmann machines for shift-invariant feature learning. In: CVRP. 2009 IEEE Computer Society Conference on CVPR, p. 2735--2742. IEEE.

\bibitem[Paiement, 2008]{Paiement2008_thesis}
Paiement, J.-F. (2008). Probabilistic Models for Music. PhD thesis, Ecole Polytechnique F\'{e}d\'{e}rale de Lausanne (EPFL).

\bibitem[Ron et~al., 1994]{Ron1994_VLMMs}
Ron, D., Singer, Y., and Tishby, N. (1994). The power of amnesia. In: Cowan, J.D., Tesauro, G., Alspector, J. (eds.) NIPS. Advances in NIPS 6, p. 176--183. Morgan Kaufmann.

\bibitem[Sutskever and Hinton, 2007]{Sutskever2007_TRBM}
Sutskever, I. and Hinton, G.E. (2007). Learning multilevel distributed representations for high-dimensional sequences. Journal of ML Research - Proceedings Track 2, 548--555.

\bibitem[Taylor and Hinton, 2009]{Taylor2009_FactoredCRBM}
Taylor, G.W. and Hinton, G.E. (2009). Factored conditional restricted {B}oltzmann machines for modeling motion style. In: Danyluk, A.P., Bottou, L., Littman, M.L. (eds.) ICML. ACM ICPS, vol. 382, p. 129. ACM.

\bibitem[Taylor et~al., 2007]{Taylor2006_CRBM}
Taylor, G.W., Hinton, G.E., and Roweis, S.T. (2007). Modeling human motion using binary latent variables. In: Sch{\"o}lkopf, B., Platt, J.C., Hoffman, T. (eds.) NIPS. Advances in NIPS 19, p. 1345--1352. MIT Press.

\bibitem[Weiland et~al., 2005]{Weiland2005_HHMM}
Weiland, M., Smaill, A., and Nelson, P. (2005). Learning musical pitch structures with hierarchical hidden {M}arkov models. Technical report, University of Edinburgh.

\bibitem[Welling et~al., 2004]{Welling_exponential_harmoniums}
Welling, M., Rosen-Zvi, M., and Hinton, G.E. (2004). Exponential family harmoniums with an application to information retrieval. In: NIPS. Advances in NIPS 17.

\bibitem[Wood et~al., 2009]{Wood09_memoizer}
Wood, F., Archambeau, C., Gasthaus, J., James, L., and Teh, Y.W. (2009). A stochastic memoizer for sequence data. In: Danyluk, A.P., Bottou, L., Littman, M.L. (eds.) ICML. ACM ICPS, vol. 382, p. 142. ACM.
}
\end{thebibliography}

\end{document}